# Improving Sentence-Level Relation Extraction Through Curriculum Learning

**Seongsik Park**    **Harksoo Kim**
Konkuk University, South Korea
{a163912, nlpdrkim}@konkuk.ac.kr

## Abstract

Sentence-level relation extraction mainly aims to classify the relation between two entities in a sentence. The sentence-level relation extraction corpus often contains data that are difficult for the model to infer or noise data. In this paper, we propose a curriculum learning-based relation extraction model that splits data by difficulty and utilizes them for learning. In the experiments with the representative sentence-level relation extraction datasets, TACRED and Re-TACRED, the proposed method obtained an F1-score of 75.0% and 91.4% respectively, which are the state-of-the-art performance.

## 1 Introduction

Relation extraction refers to the task of inferring semantic relations among entities in a text. In particular, sentence-level relation extraction mainly aims to classify the relation between two entities in a sentence. Recently, large-capacity supervised corpora, such as TACRED [1], have been released for sentence-level relation extraction. A sentence-level relation extraction corpus is constructed using distant supervision or human annotation. The corpus constructed in this manner, often contains data that are difficult for the model to infer or noise data [2]. In this paper, we propose a curriculum learning-based relation extraction model that splits data by difficulty and utilizes them for learning. In the curriculum learning process, general parameters for relation extraction are learned through easy examples and high-level inference becomes possible while difficult examples are gradually learned [3]. Classifying data according to difficulty is a highly intensive task for humans. Therefore, in this study, the difficulty classification was performed automatically using the cross review method [4].

## 2 Methodology

### 2.1 Baseline Model

Recently, most relation extraction studies used a pre-trained language[5, 6, 7] model and encoded tokens through a language model. The token encoding vector is used to calculate the entity encoding vector, and relation extraction is performed on the basis of the entity encoding vector in the sentence. The performance of sentence-level relation extraction depends on the entity marking method used in the sentence and on the entity encoding vector extraction method. We used the marking method proposed by Zhou and Chen (2021) [8] among various entity marking methods. Their method uses several punctuations as symbols for the entities and marks the entities in the sentence with the symbols. Sentences are inputted to the language model, and tokens containing each entity symbol are encoded. An entity encoding uses a symbol encoding vector that corresponds to the symbol of the entity. We also used a graph attention network (GAT) [9] using dependency graphs for more effective relationship extraction. GAT encodes syntactic structural information between two entities and predicts a suitable relation label through a bilinear classifier by concatenating the output of GAT and the encoding vector.

### 2.2 Relation Extraction Based on Curriculum Learning

Curriculum learning refers to a process in which a model starts learning from easy data and gradually learns difficult data. It mimics the human learning process Curriculum learning learns easy data quickly by finding a parameter space that is more suitable for the task and solves the problem of local minima in that space while learning difficult data. To measure the difficulty of data, we used the cross review method based on the prediction of the model. First, we split corpus into $\{0, 1, ..., n\}$ subsets by sampling without replacement. Then we created $\{0, 1, ..., n\}$ independent models trained on each subset and predicted the relation labels of the remaining subsets except the subset that was used for the training. For each data instance, $n - 1$ prediction labels were assigned and the difficulty was evaluated on the basis of the prediction results. In instances where most of the sub-models predicted the correct relation labels, it was determined that the difficulty of the data was easy and that the more the relation prediction was wrong, the more noisy or difficult the data became. Finally, we divided the entire corpus into several groups according to the difficulty of the data and allowed the model to gradually learn the entire data according to the difficulty.

| Models | P | R | F1 |
|---|---|---|---|
| SpanBERT-large [10] | 70.8 | 70.9 | 70.8 |
| KnowBERT-W [11] | 71.6 | 71.4 | 71.5 |
| BERT-MTB [12] | - | - | 71.5 |
| DeNERT-KG [13] | 71.8 | 73.1 | 72.4 |
| LUKE [14] | 70.4 | **75.1** | 72.7 |
| Typed marker [8] | - | - | 74.6 |
| Relation reduction [15] | 75.2 | 74.6 | 74.8 |
| RECENT [16] | **90.9** | 64.2 | **75.2** |
| Typed marker (our reimplementation) | 75.7 | 73.4 | 74.5 |
| Baseline (ours) | 75.8 | 74.0 | 74.9 |
| + curriculum learning (ours) | 75.3 | 74.8 | 75.0 |

Table 1: Comparison of the performances on TACRED. The bold fonts denote SOTA performance.

| Models | P | R | F1 |
|---|---|---|---|
| Entity marker | - | - | 90.5 |
| SpanBERT-large | 70.8 | 70.9 | 70.8 |
| Typed marker | - | - | 91.1 |
| Typed marker (our reimplementation) | 90.6 | 91.3 | 90.9 |
| Baseline (ours) | **91.7** | 90.7 | 91.2 |
| + curriculum learning (ours) | 91.3 | **91.5** | **91.4** |

Table 2: Comparison of the performances on Re-TACRED. The bold fonts denote SOTA performance.

## 3 Experiments

### 3.1 Implementation Details

We used the RoBERTa large model with 1,024-dimensional encodings to encode the input sentences. A dropout of 0.1 was applied to all the transformer layers and linear transformation layers. A value of 0.0005 was selected for the learning rate, and a layer decay of 0.7 was applied to RoBERTa to preserve the pre-trained weights. We used AdamW [17] to optimize the model.

### 3.2 Dataset

To evaluate the proposed model, we selected the Text Analysis Conference Relation Extraction Dataset (TACRED), which is a human-annotated relation dataset used in the Text Analysis Conference Knowledge Base Population challenge (TAC-KBP). TACRED covers 42 relation types including the *no_relation* class. In addition, we used Re-TACRED in our experiments. Although TACRED is a dataset constructed by human annotation, there are many incorrectly annotated data. Re-TACRED is a corpus refined by re-vision of the TACRED errors. Re-TACRED corrected the entity or relation type that was incorrectly tagged in TACRED and partially modified the relation type set.

### 3.3 Experimental Result

Table 1 shows the performance of the proposed model compared with the other existing models on TACRED. The typed marking model marked an entity with a special symbol and used it to extract the relation. The baseline model also applied the same entity marking method and showed a 0.4 percentage point higher performance than that of typed marking model re-implemented using the syntax graph-based GAT. When curriculum learning was applied to the baseline model, the performance improved by 0.2 percentage point. Table 2 shows the performance of the proposed model compared with the other existing models on Re-TACRED. When curriculum learning was applied to the baseline model, it showed the best performance with an F1 score of 91.4%.

## 4 Conclusion

We proposed a curriculum learning method to increase the performance of sentence-level relation extraction. The results of the experiments with TRACRED and Re-TACRED showed that the proposed method is simple but effective to increasing the performance of sentence-level relation extraction.